\documentclass[11pt,a4paper]{article}
\usepackage[hyperref]{acl2021}
\usepackage{times}
\usepackage{latexsym}
\usepackage{natbib}
\usepackage{tabularx}
\usepackage{graphicx}
\usepackage{setspace}
\usepackage{hyperref}

\usepackage{microtype}

\aclfinalcopy 
\def\aclpaperid{36} 

\usepackage{xcolor}

\title{Can Transformer Language Models Predict Psychometric Properties?}
 
\author{Antonio Laverghetta Jr., Animesh Nighojkar, Jamshidbek Mirzakhalov, \and John Licato \\
  Advancing Machine and Human Reasoning (AMHR) Lab \\
  Department of Computer Science and Engineering \\
  University of South Florida \\
  Tampa, FL, USA \\
  \texttt{\{alaverghett,anighojkar,mirzakhalov,licato\}@usf.edu} 
  }

\date{}

\begin{document}

\maketitle

\begin{abstract}
Transformer-based language models (LMs) continue to advance state-of-the-art performance on NLP benchmark tasks, including tasks designed to mimic human-inspired ``commonsense'' competencies. To better understand the degree to which LMs can be said to have certain linguistic reasoning skills, researchers are beginning to adapt the tools and concepts of the field of \textit{psychometrics}. But to what extent can the benefits flow in the other direction? I.e., can LMs be of use in predicting what the psychometric properties of test items will be when those items are given to human participants? We gather responses from numerous human participants and LMs (transformer- and non-transformer-based) on a broad diagnostic test of linguistic competencies. We then use the responses to calculate standard psychometric properties of the items in the diagnostic test, using the human responses and the LM responses separately. We then determine how well these two sets of predictions match.
We find cases in which transformer-based LMs predict psychometric properties consistently well in certain categories but consistently poorly in others, thus providing new insights into fundamental similarities and differences between human and LM reasoning.\footnote{Code and data to reproduce our experiments can be found on Github: \href{https://github.com/Advancing-Machine-Human-Reasoning-Lab/transformer-psychometrics}{https://github.com/Advancing-Machine-Human-Reasoning-Lab/transformer-psychometrics}} 
\end{abstract}

\section{Introduction}

The current generation of transformer-based language models (TLMs) \citep{vaswani2017attention} continues to surpass expectations, consistently achieving state-of-the-art results on many natural language processing (NLP) benchmark tasks.
Especially surprising is their remarkable performance on benchmark tasks designed to assess ``commonsense'' reasoning (e.g., \citealp{Wang2018,Wang2019b}), possibly owing to their ability to encode and retrieve a surprising amount of structural knowledge \cite{Goldberg2019,Hu2020a,cui2020does,petroni2019language,Davison2019}, despite initial worries that all connectionist language models in general would suffer the same limitations as previous generations \cite{Sun1992b,Sun1995,McClelland1995,Klahr1999,McLaughlin2009}. 

Understanding how TLMs reason is a complex task made more difficult by the fact that the sizes of contemporary TLMs are so large as to effectively render them black boxes. As such, researchers are continually searching for new methods to understand the strengths and limitations of TLMs. One promising approach is to draw from \textit{psychometrics}, a sub-field of psychology particularly suited to dealing with perhaps the most mysterious black box of them all: the human mind. Psychometrics is concerned with psychological measurement---i.e., how to measure latent attributes like reasoning skills, attitudes, and personality traits. Psychometricians have developed tools to measure such properties even when the mechanisms that give rise to them are not fully understood, thus suggesting a possible fruitful application of those tools to complex artificial black boxes like TLMs. Although some have called for bridging the gap between psychometrics and AI \citep{bringsjord2003artificial,bringsjord2011psychometric,bringsjord2012psychometric,DOWE201277,Hernandez2016,Wilcox2020}, the amount of work attempting to do so is limited: although some existing work attempts to use advances in psychometrics to benefit the study of TLMs, none to our knowledge have used SOTA TLMs (or even LMs in general) to benefit psychometrics.

To illustrate, assume that someone wishes to design a test to assess the degree to which a person possesses mastery of some cognitive skill $\mathcal{S}$. A good place to start is for a panel of experts to design a set of test items (questions) $\mathcal{I}$, such that they believe solving $\mathcal{I}$ requires $\mathcal{S}$. However, although many NLP benchmarks tend to consider this sufficient, the items in $\mathcal{I}$ only have \textit{face validity}, in that they only have been demonstrated to superficially test for $\mathcal{S}$. To go beyond face validity, one must assess $\mathcal{I}$'s psychometric properties by establishing their \textit{validity} (how well the items actually measure the phenomenon $\mathcal{S}$ they purport to measure), \textit{reliability} (how stable the items are as measurements), and \textit{fairness} (how well the items are free from biases against certain sub-populations of subjects).\footnote{Note however that we focus only on validity and reliability in this work.} But establishing these psychometric properties can be prohibitively costly, requiring large numbers of human participants to answer the items in $\mathcal{I}$ and iteratively refine them. This drawback motivates the central research question of our paper: \textbf{Can TLMs be used to predict psychometric properties of test items?} If so, the benefit for psychometric practitioners\footnote{In other words, professionals responsible for designing standardized tests or other evaluations meant to assess latent attributes of individuals.} is enormous, as it can reduce the need for multiple rounds of costly empirical testing. But the benefits for NLP are significant as well: knowing how the psychometric properties of items differ when applied to artificial versus human populations will give us unique insight into how they solve such problems, and how they can be improved.

\paragraph{Main Contributions of this Paper:} We present the first exploration into how well TLMs can be used to predict certain psychometric properties of linguistic test items. To do this, we identified a subset of items from the GLUE broad coverage diagnostic \citep{Wang2018}, and collected human responses on these items in order to assess simple psychometric properties, designing a novel user validation procedure to do so. We then assess the performance of 240 LMs on these diagnostic items. Our resulting analysis clearly shows that TLMs excel in modeling psychometric properties in certain sub-categories of linguistic skills, thus providing fruitful directions for future work.

\section{Related Work}
\label{sec:related}
What reason do we have to suspect that TLMs can predict the psychometric properties of test items? Although TLMs were not primarily designed to compute in a human-like way, there are some reasons to suspect that they may have the ability to effectively model at least some aspects of human linguistic reasoning: They consistently demonstrate superior performance (at least compared to other LMs) on human-inspired linguistic benchmarks \cite{Wang2018,Wang2019b}, and they are typically pre-trained using a lengthy process designed to embed deep semantic knowledge, resulting in efficient encoding of semantic relationships \cite{zhou2020evaluating,petroni2019language,Davison2019,cui2020does}. Common optimization tasks for pre-training transformers, such as the masked LM task \cite{devlin2018bert} are quite similar to the word prediction tasks that are known to predict children's performance on other linguistic skills \cite{Borovsky2012,Neuman2011,Gambi2020}. Finally, TLMs tend to outperform other LMs in recent work modeling human reading times, eye-tracking data, and other psychological and psycholinguistic phenomena \cite{merkx-frank-2021-human,Schrimpf2020a,Schrimpf2020b,hao2020probabilistic,bhatia2020transformer,Laverghetta2021, antonio2021psychometrics}.

There are many studies probing TLMs in various ways, a body of work sometimes called ``BERTology'' \cite{rogers2021primer,Belinkov2019c}. However, work explicitly bridging psychometrics with AI is less common. \citet{xue2019computational} augmented the DINA \citep{de2009dina} and DINO \citep{templin2006measurement} cognitive diagnostic models \citep{sessoms2018applications} with a feed-forward neural network that used a semi-supervised learning objective. The architecture achieved superior results to multiple baselines. \citet{ahmad2020deep} created a deep learning architecture for extracting psychometric dimensions related to healthcare, specifically numeracy, literacy, trust, anxiety, and drug experiences. Their architecture did not use transformers, and relied instead on a sophisticated combination of convolutional and recurrent layers in order to extract representations of emotions, demographics, and syntactic patterns, among others.  \citet{eisape2020cloze} examined the correlation between human and LM next-word predictions and proposed a procedure for achieving more human-like cloze probabilities. In NLP, methods from item response theory (IRT) \citep{reckase2009multidimensional} have been particularly popular. \citet{lalor2018understanding} used IRT models to study the impact of question difficulty on the performance of deep models on several NLP tasks. In a follow-up study, \citet{lalor2020dynamic} used IRT models to estimate the competence of LSTM \cite{1997LSTM} and BERT models during training. This allowed them to create a dynamic curriculum learning \cite{bengio2009curriculum} algorithm, which achieved superior performance to the same models trained using a static scheduler for several tasks. \citet{sedoc2020item} used IRT to efficiently assess chat-bots. \citet{martinez2019item} used IRT to analyze the performance of machine learning classifiers in a supervised learning task. IRT has also been used to evaluate machine translation systems \citep{otani2016irt} and speech synthesizers \citep{Oliveira2020ItemRT}, and also in computer vision \citep{richardwebster2018visual}.

This literature clearly indicates that there has been a lot of interest in applying psychometrics to AI. So far, most of this effort has focused on specific use cases, and has not attempted to broadly assess commonalities between machine and human reasoning. Most similar to our current work is  \citet{lalor2019learning}, who showed that deep models could achieve a strong correlation with IRT parameters fitted using human data on several NLP datasets. However, they compared the human responses to LSTMs and neural semantic encoders \cite{munkhdalai-yu-2017-neural-semantic}, and did not consider TLMs. Furthermore, they focused on the SNLI dataset, which is less challenging than the GLUE diagnostic and does not group questions based on fine-grained linguistic competencies.

Besides the GLUE diagnostic, other taxonomies have been proposed, such as TaxiNLI \cite{joshi2020taxinli}. Although TaxiNLI includes some types of reasoning which have no clear analogue in GLUE, many of their categories are quite similar.\footnote{Both GLUE and TanxiNLI test for temporal reasoning, but place them at different levels in the taxonomy.} Since the TaxiNLI questions were also taken from the MNLI dataset, we were concerned they would be too easy for some of the larger TLMs we planned to evaluate. We, therefore, chose to focus specifically on the challenging GLUE diagnostic set and leave TaxiNLI for future work.

\section{Gathering Language Model Data}
\label{sec:transformers}

The GLUE and SuperGLUE benchmarks \citep{Wang2018,Wang2019b} are suites of NLP tasks designed to test the general linguistic capabilities of LMs. Included as part of the GLUE benchmark is a set of diagnostic questions, called the broad coverage diagnostic, which are all formatted as natural language inference (NLI) problems. NLI problems consist of two sentences: a premise ($p$) and hypothesis ($h$), and solving such a problem involves assessing whether $p$ textually entails $h$. There are typically three choices: either $p$ does textually entail $h$ (entailment), $p$ entails that $h$ is impossible (contradiction), or $h$'s truth can not be determined from $p$ alone (neutral). The NLI task is therefore quite general and can encompass a wide variety of other ``commonsense'' reasoning tasks. The broad coverage diagnostic was manually curated by linguistics and NLP experts and is meant to assess broad psycholinguistic competencies of LMs across multiple categories. For instance, the \textit{propositional structure} category contains questions that exploit propositional logic operators; e.g., $p =$ ``The cat sat on the mat.'' and $h =$ ``The cat did not sit on the mat.'' The diagnostic covers four main categories of linguistic competencies: \textit{lexical semantics}, \textit{predicate-argument structure}, \textit{logic}, and \textit{knowledge and common sense}. These categories are further divided into multiple sub-categories, each of which covers a specific and interesting phenomenon in language. The GLUE diagnostic thus aims to be a comprehensive test of linguistic reasoning skills, making it suitable for our present study.

To evaluate our models, we selected a subset of the GLUE diagnostic questions that were a member of only one sub-category, to better isolate factors. In most cases, there were enough questions in a single sub-category that we could just drop all questions that belonged to multiple sub-categories, further details on this preprocessing can be found in Appendix \ref{append:LM experiments}. After performing preprocessing, we had 811 remaining diagnostic questions encompassing 20 sub-categories. Each sub-category had at least 15 questions, and we selected 7 of the sub-categories to use in our experiments:

\begin{enumerate}
    \item \textit{morphological negation} (MN)
    \item \textit{prepositional phrases} (PP)
    \item \textit{lexical entailment} (LE)
    \item \textit{quantifiers} (Q)
    \item \textit{propositional structure} (PS)
    \item \textit{richer logical structure} (RLS)
    \item \textit{world knowledge} (WK)
\end{enumerate}

We selected these 7 sub-categories based on how much the average performance of the LMs improved after pre-training and finetuning. A substantial performance improvement indicated the category was solvable by the models, and would therefore provide a meaningful comparsion to the human data. We gathered responses to the diagnostic from a wide array of TLMs, including BERT \citep{devlin2018bert}, RoBERTa \citep{liu2019roberta}, T5 \citep{raffel2020exploring}, ALBERT \citep{lan2020albert}, XLNet \citep{yang2019xlnet}, ELECTRA \citep{clark2020electra}, Longformer \citep{beltagy2020longformer}, SpanBERT \citep{joshi2019spanbert}, DeBERTa \citep{he2020deberta}, and ConvBERT \citep{jiang2020convbert}. Each of these models differs from the others along one or more factors, including underlying architecture, pre-training objective and data, or the general category the model belongs to (autoregressive, autoencoding, or sequence-to-sequence). For most of these models we used the Transformers \citep{wolf-etal-2020-transformers} implementation, the exception being T5, which was implemented in PyTorch Lightning \citep{falcon2019pytorch}. We use LSTM-based LMs \citep{1997LSTM} as a baseline, further details on the LMs can be found in Appendix \ref{append:LM experiments}.

We used the SNLI \citep{snli:emnlp2015}, MNLI \citep{N18-1101}, and ANLI \citep{nie-etal-2020-adversarial} training and dev sets to finetune our models. We found that the amount of finetuning data had a significant impact on final diagnostic performance. Therefore, to increase the variance in our results as much as possible we used the following training set partitions for all model configurations: 


\begin{itemize}
    \item SNLI alone
    \item MNLI alone
    \item SNLI + MNLI
    \item SNLI + MNLI + ANLI
\end{itemize}

Both the train and dev sets were shuffled before every trial. We finetuned our models for between 5 to 10 epochs. We used the reported Matthews correlation \citep{matthews1975comparison} on the dev set during training to determine when the performance had saturated; when this correlation stopped consistently increasing for at least a few dev set evaluations we stopped training. We evaluated on the dev set every 15,000 steps. All the transformer's key hyperparameters were selected in a similar way to the study by \citet{lalor2019learning}. For all models, we used a learning rate of $1*10^{-5}$ and a max sequence length of 175. Since running even a small grid search to optimize the hyperparameters of each model would have dramatically increased the number of trials, we instead chose to fix these hyperparameters to be similar to what was used in prior work (e.g. \citealp{devlin2018bert}). We also found that nearly all models consistently achieved a Matthews correlation of about 0.5 or higher on the dev set, and thus concluded that our hyperparameters were suitable. It is important to note that our goal in finetuning was not to completely optimize the model's performance on these NLI datasets. Rather, since the diagnostic is formatted as an NLI task, we hoped that finetuning would help the models to learn what the output labels should be.\footnote{Finetuning T5 is necessary to avoid random output.} To evaluate these models, we experimented with four different training regimes:

\begin{itemize}
    \item \textbf{Zero shot:} The model is initialized with random weights in the hidden layers and is evaluated on the diagnostic without any training. This is meant to test whether there is any property of the architecture itself which is useful for solving the diagnostic.
    \item \textbf{Pre-train, no finetune:} The model is pre-trained but not finetuned.
    \item \textbf{No pre-train, finetune:} The model weights are initialized randomly, but we finetune the model before evaluating it.
    \item \textbf{Pre-train and finetune:} The model is pre-trained and finetuned.
\end{itemize}

For BERT, we experimented with both \citeauthor{devlin2018bert}'s pre-trained models, and a BERT model we trained from scratch. Our BERT model had an identical architecture to \textit{bert-base} and was pre-trained on Google's One Billion Words corpus \citep{chelba2014one}. We used the same hyperparameters from the BERT paper \cite{devlin2018bert}, using a learning rate of $4*10^{-5}$, a max sequence length of 128, a warmup ratio of 0.01, and a weight decay of 0.01. We used the Transformers library to pre-train this model, and saved every end-of-epoch checkpoint. We pre-trained for 52 epochs and used every 10th checkpoint to gather diagnostic data separately. This allowed us to study the effect pre-training had on diagnostic performance. 

In summary, this process allowed us to vary the underlying architecture, the number of trainable parameters, and the amount of finetuning data used in each trial. This allowed us to treat each trained model as effectively being a different ``individual'' (and we will refer to them as such), which might have a radically different cognitive profile from its counterparts. For example, a \textit{roberta-base} model that was pre-trained and finetuned on all 3 NLI datasets might produce very different response patterns than a \textit{roberta-large} model evaluated zero-shot. We used three Tesla V100 GPUs with 32GB of video memory each, as well as preemptable GPUs on Google Colab,\footnote{\hyperlink{https://colab.research.google.com}{https://colab.research.google.com}} to train all models. Wherever possible, we used Apex\footnote{\hyperlink{https://github.com/NVIDIA/apex}{https://github.com/NVIDIA/apex}} to speed up training.

\section{Human Studies}
\label{sec:humans}
As our purpose in gathering this LM data was to evaluate it against human performance, we additionally ran a human study. To do this, we recruited workers on Amazon Mechanical Turk (mTurk\footnote{\hyperlink{https://www.mturk.com}{https://www.mturk.com}}) to complete our subset of GLUE diagnostic questions. While mTurk makes conducting large-scale human studies convenient, there are also well-documented problems with participants not completing tasks in good faith \citep{berinsky2014separating,berinsky2016can,keith2017systems}. There are multiple techniques for filtering out bad-faith participants, such as the use of ``attention check'' questions, sometimes called ``instructional manipulation checks'' \cite{hauser2015it}, which are designed so that a good-faith participant would be unlikely to get them incorrect. But this alone would not suffice for our purposes here, as we want a certain amount of low-scoring participants on some sub-categories, so that the population variances on sub-category items would better reflect their actual variances.\footnote{If we only kept high-performing participants, the item variances would be skewed to be low and roughly the same, which would not reflect the true variances we would expect to see from a large population of good-faith participants.} Therefore, we designed a procedure for distinguishing bad-faith from low-performing participants.

We first obtained attention checks from the ChaosNLI dataset \citep{nie2020what}, which gathered over 450,000 human annotations on questions from SNLI and MNLI. Since each question in ChaosNLI was annotated by 100 different workers, if inter-annotator agreement for a given question is high, we conclude that question is likely extremely easy to solve. These questions were also in the same format as the diagnostic questions, which made it less likely that workers would realize they were being given an attention check. We gathered 36 questions from ChaosNLI where the agreement for the correct label was at least 90\%. The labels for this subset were perfectly balanced. These were enough questions to ensure that each phase of our trials used a unique set of attention check questions.

The human studies were split up into 5 phases, and workers who did sufficiently well in a given phase were given a qualification to continue to the next phase:

\begin{enumerate}
\item \textbf{On-boarding:} A qualifying HIT (human intelligence task) open to any worker located in the United States, who had completed at least 50 HITs with an approval rating of at least 90\%. The HIT consisted of 5 attention check questions, given to each worker in the same order. We gathered responses from up to 200 workers.\\
\item \textbf{Phase 1:} Included questions from \textit{morphological negation}, and 3 attention checks. We gathered up to 45 responses.\\
\item \textbf{Phase 2:} Included questions from \textit{lexical entailment} and \textit{prepositional phrases}, as well as 6 attention checks. We gathered up to 36 responses.\\
\item \textbf{Phase 3:} Included questions from \textit{quantifiers} and \textit{propositional structure}, as well as 6 attention checks. We gathered up to 27 responses.\\
\item \textbf{Phase 4:} Included questions from \textit{richer logical structure} and \textit{world knowledge}, as well as 6 attention checks. We gathered responses from all accepted workers from Phase 3.
\end{enumerate}

In each phase, questions were randomly ordered, except for attention checks which were spread evenly throughout the survey. We used Qualtrics\footnote{\hyperlink{https://www.qualtrics.com}{https://www.qualtrics.com}} to create the surveys for each HIT and collect the responses. Participants were first presented with instructions for the task and some examples, which were based on the instructions originally given to annotators for the MNLI dataset.\footnote{\hyperlink{https://nyu-mll.github.io/GLUE-human-performance/mnli.html}{https://nyu-mll.github.io/GLUE-human-performance/mnli.html}} The questions from each category were a randomly chosen subset of 15 questions tested on the LMs for that category, balanced for each label. For each question, workers also had to provide a short justification statement on why they believed their answer was correct, which was used to help filter out bad faith participants. To validate the responses to our surveys, we developed the following authentication procedure:

\paragraph{Stage 1:} Look for duplicate IPs or worker IDs, indicating that the worker took the HIT more than once. If there are any, reject the second and future HITs, but keep the first submission.\\
\textbf{Stage 2:} If the worker's overall score was less than 40\%, reject the HIT. If their overall score was greater than 60\%, accept the HIT. For workers who scored between 40\% and 60\%, we still rejected the HIT if they got less than 75\% of the attention checks correct. \\
\textbf{Stage 3:} Finally, examine the justifications of all workers not previously rejected. Here we were looking for simple, but clear, reasons for why workers chose their answer. We included this step because we found in a pilot study that workers sometimes provided nonsensical justifications for their answers even when they did well on the survey, making it unclear whether they were truly paying attention. We checked that the justifications appeared relevant to the question (some workers seemed to paste random text from other websites into the justification), that they did not paste part of the question for their justification, that they did not use the same justification for every question, and that they did not use short nonsensical phrases for their justification (some workers simply wrote ``good'' or ``nice'' as their justification). This allowed us to keep some low-scoring participants who had put genuine effort into the task.

Manual inspection of the resulting responses suggested that workers who passed stage 3 consistently gave higher quality responses than those who did not. These workers gave more detailed justifications that clearly articulated their thought process, often citing specific details from the question. On the other hand, workers who failed to give good justifications also tended to perform quite poorly, generally scoring at or below random chance, which further indicated that they were not actually paying attention. We, therefore, believe the use of justifications helped us gather higher-quality responses. Further details on the human study can be found in Appendix \ref{append:humans}.

\section{Experimental Results}
\label{sec:results}
Using the procedures described in \S \ref{sec:transformers} and \S \ref{sec:humans}, we gathered results from 27 human participants and 240 neural LMs (183 transformer-based and 57 LSTM-based). In addition to the LSTMs, we also include a true random baseline which simply guesses randomly on every question. In the following experiments, we use the human performance on each category as the basis for analyzing the performance of the artificial populations, specifically in terms of how well each artificial population's responses correlate with the human data.

\begin{table}[htb]
\centering
\singlespacing
\scalebox{0.75}{
\begin{tabular}{c|c|c|c}
\textbf{Category}                 & $\mathbf{D_T}$         & $\mathbf{D_L}$        & $\mathbf{D_R}$         \\
\hline
\textbf{MN}   & \textbf{-0.28, \textless{}0.5}     & 0.27, \textgreater{}0.5    & -0.14, \textgreater{}0.5              \\
\textbf{PP}    & \textbf{0.86, \textless{}0.001} & 0.47, \textless{}0.1  & 0.42, \textless{}0.5  \\
\textbf{LE}       & \textbf{0.62, \textless{}0.05}  & 0.17, \textgreater{}0.5  & -0.22, \textless{}0.5  \\
\textbf{Q}              & \textbf{0.57, \textless{}0.05}   & -0.22, \textless{}0.5 & 0.41, \textless{}0.5     \\
\textbf{PS}  & \textbf{0.93, \textless{}0.001} & 0.27, \textless{}0.5 & 0.37, \textless{}0.5  \\
\textbf{RLS} & 0.28, \textless{}0.5   & -0.03, \textgreater{}0.5  & \textbf{-0.37, \textless{}0.5} \\
\textbf{WK}          & \textbf{0.79, \textless{}0.001} & 0.46, \textless{}0.1   & -0.25, \textless{}0.5   
\end{tabular}
}
\caption{Given $D_H$, Spearman correlation and p-values were calculated with transformer-based ($D_T$), LSTM-based ($D_L$), and random ($D_R$) estimates of problem difficulty (percentage of the population that got the item correct). Note here we have bolded cells whose correlations (absolute values) were highest, but their p-values were not always significant.}
\label{tab:difficulties}
\end{table}

\subsection{Classical Test Theory}
We began by examining how well TLMs could predict simple problem difficulty in the human data. This measure comes from classical test theory and is calculated simply as how many members of the population get a given item right. For each item $i$ in a given sub-category in our subset of the GLUE diagnostic, we calculated the percentage of human participants who got that question correct ($D^{i}_{H}$), and then the corresponding percentage for the TLMs ($D^{i}_{T}$), LSTM-based LMs ($D^{i}_{L}$), and the random baseline ($D^{i}_{R}$). We then calculated the Spearman correlation \citep{spearman1961proof} between $D^{i}_{H}$ and each of the other populations. Results are shown in Table \ref{tab:difficulties}. In almost all cases, TLMs achieve a much stronger correlation with the human data than either baseline, and most were statistically significant. The main exceptions are \textit{morphological negation} (MN) and \textit{richer logical structure} (RLS), both of which fail to produce strong correlations. As we will see, this pattern will repeat in other measurements as well. 

\paragraph{IIC-based Clustering} An important idea in psychometrics is that questions that rely on the same skills should have similar chances of being answered correctly by a given participant \cite{rust2014modern}. Whether questions rely on similar skills can be tested using the inter-item correlation (IIC) between two items, where high IIC suggests that the items rely on similar underlying reasoning skills. Thus, it can be assumed that if items cluster together when using IIC as a distance metric, they rely on similar underlying cognitive skills. To explore this, given a correlation measure $c$ ranging from -1 to 1, we convert it into a distance metric by taking $1-c$. We use this metric to apply k-medoids clustering to the diagnostic questions, using the silhouette method \citep{rousseeuw1987silhouettes} to find the optimal number of clusters. For each sub-category, we perform clustering using human, transformer, LSTM, and random data separately ($H$, $T$, $L$, and $R$ respectively). We use the k-medoids implementation from scikit-learn extra\footnote{\hyperlink{https://github.com/scikit-learn-contrib/scikit-learn-extra}{https://github.com/scikit-learn-contrib/scikit-learn-extra}} and use scikit-learn \citep{scikit-learn} to calculate the silhouette coefficient.

After clustering, for each pair of items ($i,j$) we define  $C^{D}_{i,j}$ as $1$ if $i$ and $j$ are in the same cluster as determined by dataset $D$ $\in \{H,T,L,R\}$. Finally, to determine how well clusters from the LM responses match the human responses, we calculate Pearson correlation \citep{pearson1895notes}
between $C^H$ and each of $C^T$, $C^L$, and $C^R$. Results are shown in Table \ref{tab:clusters}. Similar to Table \ref{tab:difficulties}, we see statistically significant correlations from TLMs in every sub-category, except for \textit{morphological negation} (MN), where TLMs again achieve only weak correlation.

\begin{figure}
 \centering
 \includegraphics[width=1\linewidth]{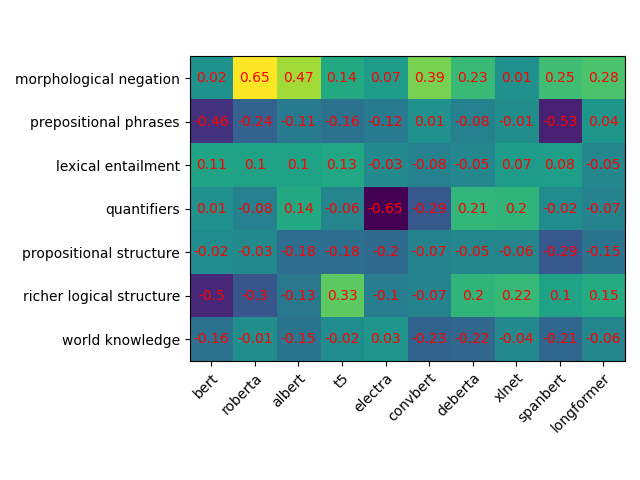}
 \caption{Change in correlation for each TLM architecture on each category, compared to the entire TLM population. Best viewed in color.}
 \label{fig:spdAblation}
\end{figure}

\paragraph{Per Model Analysis} The previous results give us some insights into the performance of the entire TLM population. However, individual transformers might differ somewhat in the specific skills they are proficient in. To study this, we performed the same simple problem difficulty experiment, but this time only used the diagnostic results from a single transformer architecture (for instance just BERT). We did this for each architecture, and then on each diagnostic sub-category, we computed the difference between the single architecture's correlation and the overall correlation from Table \ref{tab:difficulties}. The heatmap in Figure \ref{fig:spdAblation} shows the results, with cooler colors indicating a stronger decrease in correlation and warmer colors indicating a stronger increase. In many cases, the correlation is almost the same as the value reported in Table \ref{tab:difficulties}. However, in some cases the difference is striking. For example, RoBERTa gets a much stronger correlation on \textit{morphological negation} than any other model. Overall, it appears that most models are achieving close to the mean correlation, but there are a few significant differences.

\begin{table}[htb]
\centering
\singlespacing
\scalebox{0.75}{
\begin{tabular}{c|c|c|c}
\textbf{Category}                 & $\mathbf{C_T}$         & $\mathbf{C_L}$        & $\mathbf{C_R}$         \\
\hline
\textbf{MN}   & 0.18, \textless{}0.1   & \textbf{0.40, \textless{}0.001}    & -0.14, \textless{}0.5    \\
\textbf{PP}    & \textbf{0.31, \textless{}0.01}  & -0.15, \textless{}0.5    & -0.01, \textgreater{}0.5 \\
\textbf{LE}       & \textbf{0.31, \textless{}0.01} & -0.03, \textgreater{}0.5 & -0.16, \textless{}0.5    \\
\textbf{Q}              & \textbf{0.24, \textless{}0.05}  & -0.01, \textgreater{}0.5 & 0.06, \textgreater{}0.5  \\
\textbf{PS}  & \textbf{0.51, \textless{}0.001} & 0.03, \textgreater{}0.5  & 0.04, \textgreater{}0.5  \\
\textbf{RLS} & \textbf{0.46, \textless{}0.001} & -0.07, \textless{}0.5  & 0.04, \textgreater{}0.5  \\
\textbf{WK}          & \textbf{0.28, \textless{}0.01} & 0.00, \textgreater{}0.5  & -0.09, \textless{}0.5   
\end{tabular}
}
\caption{Pearson correlation and p-values for how well items clustered using human responses match the clusters which used transformer-based ($C_T$), LSTM-based ($C_L$), and random ($C_R$) items.}
\label{tab:clusters}
\end{table}

\subsection{Item Response Theory}
Models from classical test theory have an important shortcoming: they provide no clear way to separate the characteristics of the test taker and the test items. In practice, the observed performance on a test is affected by both the test taker and the test itself. This intuition is formalized in a psychometrics approach known as item response theory (IRT), in which both item characteristics and individual ability are modeled and used to predict performance \citep{baker2004item}. IRT models are often regarded as more informative than classical models and have become standard tools when designing evaluation scales. Formally, let $j$ be an individual taking a test, $i$ be an item on that test, and $\theta_{j}$ be that individual's latent ability. Then the probability that $j$ answers $i$ correctly is defined as:

\begin{equation}
    P(y_{i} = 1 |\theta_{j}) = c_{i} + \frac{1 - c_{i}}{1 + e^{-a_{i}(\theta_{j} - b_{i})}}
\end{equation}

\noindent Where $a_{i}$, $b_{i}$, and $c_{i}$ are \textit{item parameters} and $y_{i} = 1$ indicates a correct answer. $a_i$ is the discrimination parameter, which refers to how effective the item is for picking out high versus low ability test takers. $b_i$ is the difficulty parameter, which models how easy or difficult the item is. Finally, $c_i$ is the probability of guessing correctly. If both guessing and discrimination are held constant, we get the one-parameter or Rasch model \citep{rasch1993probabilistic}. Given a large number of human responses to a set of items, parameters for IRT models can be estimated using the marginal maximum likelihood method and expectation maximization \citep{bock1981marginal}.

Since TLMs correlated well with humans using the classical techniques we tested, we wished to examine whether this would still hold using IRT models. To do this, we used the diagnostic results from each population to fit Rasch models. We used the ltm R package to fit all models \citep{rizopoulos2006ltm}. This gave us separate difficulty parameter estimates $b_i$ for each item $i$, for each population. To determine how well the difficulty parameters matched between populations, we calculated the Pearson correlation between the $b_i$ using our human response data ($H$), and the $b_i$ obtained using the other populations ($T$, $L$, $R$). Results are shown in Table \ref{tab:rasch}. As before, TLMs consistently get a stronger correlation than either baseline on most sub-categories, except for \textit{morphological negation} (MN) and \textit{richer logical structure} (RLS). Interestingly, LSTM-based LMs achieved statistically significant and stronger correlations than TLMs on certain sub-categories: \textit{world knowledge} (WK) and \textit{prepositional phrases} (PP). The only other experiment where LSTM-based LMs achieved stronger correlation was reported in Table \ref{tab:clusters}, where they achieved superior correlation to TLMs on \textit{morphological negation} (MN).

\begin{table}[htb]
\centering
\singlespacing
\scalebox{0.75}{
\begin{tabular}{c|c|c|c}
\textbf{Category}                 & $\mathbf{D_T}$         & $\mathbf{D_L}$        & $\mathbf{D_R}$         \\
\hline
\textbf{MN}   & 0.08, \textgreater{}0.5     & \textbf{0.29, \textless{}0.5}    & 0.19, \textgreater{}0.5              \\
\textbf{PP}    & 0.48, \textless{}0.1 & \textbf{0.69, \textless{}0.01}  & -0.25, \textless{}0.5  \\
\textbf{LE}       & \textbf{0.88, \textless{}0.001}  & -0.06, \textgreater{}0.5  & 0.14, \textgreater{}0.5  \\
\textbf{Q}              & \textbf{0.61, \textless{}0.05}   & 0.03, \textgreater{}0.5 & 0.12, \textgreater{}0.5     \\
\textbf{PS}  & \textbf{0.61, \textless{}0.05} & 0.05, \textgreater{}0.5 & -0.25, \textless{}0.5  \\
\textbf{RLS} & 0.16, \textgreater{}0.5   & -0.05, \textgreater{}0.5  & \textbf{-0.31, \textless{}0.5} \\
\textbf{WK}          & 0.52, \textless{}0.05 & \textbf{0.59, \textless{}0.05}   & -0.1, \textgreater{}0.5   
\end{tabular}
}
\caption{Pearson correlation and p-values for transformer-based ($D_T$), LSTM-based ($D_L$), and random ($D_R$) estimates of problem difficulty computed using Rasch models.}
\label{tab:rasch}
\end{table}

\section{Discussion}
\label{sec:discuss}
Our analysis has revealed some interesting patterns that would have been difficult to discern using traditional evaluation metrics. Overall, TLMs perform consistently better than either of our baselines in modeling human psychometric properties. However, this improvement is also not uniform across all psycholinguistic categories. In fact, we have found some regularities in this regard. For instance, TLMs failed to achieve a strong correlation on \textit{morphological negation} in all cases. This might be explained by two facts: there is little relative variance in the human responses in this sub-category, and the average accuracy of human participants was above 90\%, as opposed to LM accuracy of 55\%. This sub-category also tests for reasoning over negation, which prior studies found that transformers struggle with \citep{rogers2021primer}. This ability to analyze the specific kinds of reasoning transformers have become proficient in is a clear advantage psychometrics have over typical NLP evaluations. The NLP community is becoming increasingly aware of the need to construct more fine-grained evaluation benchmarks \cite{Wang2018,joshi2020taxinli}, and we believe our work complements these efforts nicely.

Of course, this study also has limitations. The number of human participants in our study was somewhat small compared to typical psychometrics studies, which makes it difficult to draw stronger conclusions. One of the main criticisms IRT models draw is that they can require thousands of responses to get good estimates of the latent parameters \cite{MIN2021100963}. As stated earlier, practical limitations on population size is a common problem in psychometrics research, one which our present work hopes to alleviate somewhat. Future work will need to repeat our experiments with much larger population sizes, and also take measures to ensure sufficient diversity in the study population (e.g., age, income, education level, English fluency, etc.). Improvements in the computational efficiency of TLMs is likely also necessary for our approach to be practical, as it is unlikely most pyschometricians have access to extensive GPU resources. One possible solution would be to identify a subset of TLMs that preserves the psychometric properties of the entire population, which might allow us to achieve similar results with fewer models.

Furthermore, although we reported in detail on certain psychometrics measures where our method demonstrated promising results for TLMs, it is worth reporting that certain other measures we examined did not appear to align well. For example, item-total correlations using human data did not appear to correlate with any LM data better than with the random baseline. Likewise, our LMs failed to predict average inter-item correlations between either random subsets of items or our diagnostic sub-categories. More work is needed to better understand why. 

Finally, while our experiments have given us some insights into the validity and reliability of the diagnostic items, it is unclear whether our approach can allow us to measure their fairness. Although it is an important property, fairness is somewhat more controversial than other psychometric properties, in part because there are multiple interpretations of what constitutes test bias \cite{warne2014exploring}. Being able to probe the fairness of items would have interesting downstream applications. For instance, it might indicate whether a diagnostic gives an unfair advantage to certain types of classifiers.

\section{Conclusion}
We believe our work offers a clear path forward for bridging psychometrics and AI. The use of psychometrics measures gives us a more nuanced understanding of the latent abilities of LMs than single-valued measures like accuracy or $F_1$ can provide. Furthermore, the increasingly powerful ability of TLMs to model human ``commonsense" reasoning and knowledge suggests new ways to predict psychometrics properties of test items, reducing the need for costly human empirical data.

\section*{Acknowledgments}
This material is based upon work supported by the Air Force Office of Scientific Research under award numbers FA9550-17-1-0191 and FA9550-18-1-0052. Any opinions, findings, and conclusions or recommendations expressed in this material are those of the authors and do not necessarily reflect the views of the United States Air Force.


\appendix
\section{Additional Details on Language Model Experiments}
\label{append:LM experiments}
To create the subset of the GLUE diagnostic, there were three cases where we needed to merge members of one sub-category into another to prevent overlap:

\begin{enumerate}
    \item \textit{negation} and \textit{double negation} questions were merged into \textit{morphological negation}.
    \item \textit{symmetry/collectivity} was merged into \textit{core arguments}.
    \item Questions in both \textit{world knowledge} and \textit{named entities} were merged into \textit{named entities}.
\end{enumerate}

Each of these was cases where the sub-categories overlapped highly. For a full listing of the sub-categories and their descriptions, see \cite{Wang2018}. We experimented with multiple different snapshots of each TLM, which differed in the number of trainable parameters. We obtained these snapshots from HuggingFace.\footnote{\hyperlink{https://huggingface.co/models}{https://huggingface.co/models}} For each model we used a smaller version, designated with the \textit{small} or \textit{base} suffix, and a larger version, designated with the \textit{base} or \textit{large} suffix. For example, for BERT we experimented with both \textit{bert-base} and \textit{bert-large}, where \textit{bert-large} had more trainable parameters. For ALBERT, we used the \textit{base} and \textit{xxlarge} versions.

For the LSTMs, we used a PyTorch implementation designed specifically for NLI.\footnote{\hyperlink{https://github.com/pytorch/examples/tree/master/snli}{https://github.com/pytorch/examples/tree/master/snli}}  We initialized the LSTM-based LMs with GloVe word embeddings \citep{pennington2014glove}. We ran a non-exhaustive grid search to generate a population of LSTMs, changing the number of recurrent layers, size of the hidden layers, learning rate, and dropout \citep{srivastava2014dropout} probability.

\section{Human Study Details}
\label{append:humans}

We paid workers the following amount for each phase:

\begin{itemize}
    \item \textbf{On-boarding: } \$0.50
    \item \textbf{Phase 1: } \$3.60
    \item \textbf{Phase 2: } \$7.20
    \item \textbf{Phase 3: } \$7.20
    \item \textbf{Phase 4: } \$7.20
\end{itemize}

Our payment structure was designed to incentivize workers to put forth their best effort when completing the task. Workers were informed that successfully completing each task would award them the opportunity to earn additional payment on each subsequent phase. However, if on a given phase a worker failed our authentication protocol we rejected their work and did not pay them. Workers were informed before starting every study that we would evaluate the quality of their work, and that it might be rejected if we found evidence that they did not put forth an honest effort.

\end{document}